# Adaptive Planning Search Algorithm
# for Analog Circuit Verification

C. Manolache*, C. Andronache*, A. Caranica*, H. Cucu*,
A. Buzo**, C. Diaconu**, G.Pelz**

*University Politehnica of Bucharest
*E-mail: cristian.manolache@upb.ro
**Infineon Technologies
**E-mail: andi.buzo@infineon.com

*Abstract* — *Integrated circuit verification has gathered considerable interest in recent times. Since these circuits keep growing in complexity year by year, pre-Silicon (pre-SI) verification becomes ever more important, in order to ensure proper functionality. Thus, in order to reduce the time needed for manually verifying ICs, we propose a machine learning (ML) approach, which uses less simulations. This method relies on an initial evaluation set of operating condition configurations (OCCs), in order to train Gaussian process (GP) surrogate models. By using surrogate models, we can propose further, more difficult OCCs. Repeating this procedure for several iterations has shown better GP estimation of the circuit's responses, on both synthetic and real circuits, resulting in a better chance of finding the worst case, or even failures, for certain circuit responses. Thus, we show that the proposed approach is able to provide OCCs closer to the specifications for all circuits and identify a failure (specification violation) for one of the responses of a real circuit.*

## 1. Introduction

Pre-silicon (pre-SI) verification is one of the most time-consuming phases during circuit development. It can range anywhere from 50% [1] to 70% [2] of the time assigned to circuit design. Thus, efficiently automating the pre-SI verification is crucial for achieving optimal and faster circuit development.

The main problem in circuit verification is the impossibility of exploring every input OCC. Even if we consider only a coarse level of discretization of each OC (e.g., minimum, nominal, and maximum value) and factoring in that the operating conditions can easily be in the tens, a full factorial test is not feasible. Therefore, in developing circuits, a more effective exploration technique is necessary.

In practice, there are several methods of sampling the input hyperspace. A relatively rudimentary approach is to subsample the space with Monte Carlo simulations [3]. Advanced techniques involve the use of machine learning methods [4] [5], enhancing already sophisticated deterministic algorithms (such as Symbolic Quick Error Detection) [6], or even utilizing FPGA prototyping [2].

In this paper we propose a machine learning method to identify specification violation for a certain circuit, with minimal simulation budget requirements. Building upon our previous work in [7] and [8], we introduce a new phase named Adaptive Planning (AP). This new step uses the Gaussian Process model in order to iteratively propose better candidates. To evaluate the effectiveness of the algorithm improvement, we assess the technique on both synthetic and real circuits. Our work reports on crucially improving the algorithm, which can now identify OCCs that lead to specification violations for one of the real circuits. This is very important provided that the previously proposed methods [7][8] were not able to cover these OCCs.

## 2. Method

In this paper, we continue our previous work [7] [8], with further enhancements to the candidate selection algorithm. The main goal of the algorithm is to find candidates (i.e., OCCs) that output responses which might violate the circuit's specifications. A summary of the prior version of the method and details on the improvements are presented in the following subsections.

### A. Prior method overview

In our previous work, we trained GP models on an initial set of data, denoted Fixed Planning (FP) data. The FP data represented a combination of Orthogonal Arrays (OA) and Latin Hypercube Samples (LHS), in the OC hyperspace. Based on an evaluation set of candidates, we used Gradient Descent (GD), together with the estimates of the GP surrogate models, in order to obtain a better pool of potential candidates. From this pool, we select the candidates which might lead to obtaining better worst cases. The initial OC evaluation set is obtained from a $L^{\wedge}OC$ full-factorial (FF) number of samples, to which we added $N$ LHS samples, where $L$ is the number of levels (2 in our case), and $N$ is the difference between 5000 and the total FF samples. After we obtain the new candidate pool, we use the GP (one trained for each response of the circuit) to extract one OCC, as the next candidate to be simulated. This selection is made based on the standard Lower Confidence Bound (LCB) score. The lowest score would indicate the highest chance to fail the specification.

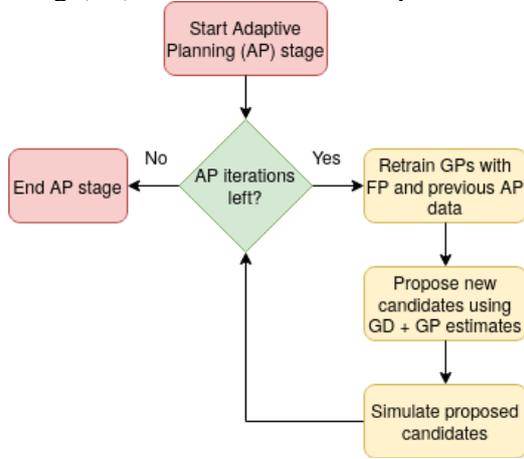

**Fig. 1** Adaptive planning stage diagram

### B. Methodology updates

In our previous method, the GPs only made one OCC selection for each response. In order to address this limitation, we introduced a new stage in our algorithm, illustrated in Fig. 1, named Adaptive Planning (AP). In this new stage, we retrain the GPs for several iterations, with samples from FP and all the candidates simulated in previous iterations. Through this process, we obtain better GPs which will better model the circuit responses. Thus, the GPs progressively propose candidates that are closer to the worst cases, therefore gradually taking closer steps to the OCC that might prove circuit failure.

Another new and important aspect of the circuits used for testing our new-stage algorithm is that they contain Process Corners (PCs). PCs represent process variations of the P-MOS, N-MOS transistors varied during manufacturing. These variations slightly change the intended nominal behavior. As opposed to OCs, which have a continuous range of values, PCs can only take categorical values (e.g., "fast", "slow", "nominal"). We encode each PC value in our experiments using two integers. This encoding was motivated by the employed Bayesian optimization library: BoTorch [9]. The six PC values (slow gain, slow, nominal, slow fast, fast slow and, fast) will be represented by two dimensions, where each dimension can take an integer value, based on a set of valid corner combinations for the given circuit. Therefore, PCs add another layer of complexity to the problem. Our experiments involve circuits with a single PC that can take one of six different values. These circuits serve as a suitable testbed for evaluating the performance of our algorithm, specifically its ability to handle circuits with process corners.

## 3. Circuits

### A. Real circuit

In our previous work, we evaluated the proposed algorithm on a synthetic circuit. In this work, a real circuit has also been introduced for extended evaluation. The circuit, denoted L2, is a voltage regulator used for various components that require a stable voltage, such as microcontrollers and DC-DC converters. The circuit has 27 design variables (which were fixed by the designer during circuit sizing), 7 OCs and 1 PC which we vary during pre-SI verification and 3 responses: gain margin (GM), phase margin (PM), and power supply rejection ratio (PSRR). The responses must meet certain specifications regarding the OCC: GM > 8, PM > 20 and PSRR > –26.

### B. Synthetic circuit

In order to speed up development and algorithm evaluation, we opted to also use synthetic circuits. These have the advantage of offering fast response values, as opposed to real circuits, that take a long time to simulate. We used a variant of the previously used synthetic circuit [7], where we introduced the notion of process corners. This allowed us to get an intuition of the algorithm's performance, before evaluating on a real circuit. One important aspect regarding synthetic circuits is that we know the minimum and maximum responses, which allows us to make a proper evaluation of the algorithm, as opposed to real circuits where these values are unknown.

In developing the synthetic circuit, the Infineon engineers drew upon their extensive experience in order to test on a relevant circuit variant. The circuit is the one described in [7] with the addition of PC according to the following formula:

$$R = aR + b \qquad (1)$$

, where $a$ and $b$ are coefficients with different values depending on the PC value, and $R$ denotes the circuit responses in [7].

## 4. Experimental results

### A. Baseline

This section presents our baseline results for the synthetic circuit and the L2 circuit. All experiments in this section have been performed on 10 random seeds, in order to reduce the impact of randomness. The FP stage consisted of 100 samples. As evaluation metrics, we have opted to use the metrics from our previous work, namely Average Relative Value Error (ARVE) and Maximum Relative Value Error (MRVE) [7]. While these metrics can be employed to assess the performance of the algorithm on synthetic circuits, this approach is not applicable to real circuits, where the true minimum and maximum are unknown. For this evaluation, the outcomes will be assessed in terms of absolute values of the responses. For the synthetic circuit, we have presented the results in Fig. 2 (FP and AP 1 rows), where we can see the ARVE and MRVE values for all 10 seeds, as well as the average. In terms of ARVE, our baseline results show an improvement from FP stage to AP iteration 1 from 2.07% to 0.75% (yellow and gray). Similarly, we can see an improvement in terms of MRVE from 6.49% to 2.87% (red and green).

Since we cannot use the same evaluation metrics for the L2 circuit, we compare the results in terms of absolute values for the 3 responses. The results are highlighted in Table 1, where we can see the minimum, maximum, and average values across all 10 seeds. In order to gain insight into the range of values that can be obtained with different seeds, the minimum and maximum values are reported. This highlights the impact of randomness and provides a sense of how far apart values can be from each other. In the case of the L2 circuit, in terms of average results, AP iteration 1 does not obtain better results to our FP stage. One possible cause would be that the L2 circuit is more complex in terms of responses and that the GPs did not have enough training data in order to have a good estimate of the responses.

### B. Algorithm update

In this subsection, we present the results obtained after Adaptive Planning. For the synthetic circuit, we have a set number of 10 iterations (results presented in Fig. 2). Here, we can see an improvement in terms of both ARVE and MRVE, from 0.75% to 0.35% (gray and black) and from 2.87% to 1.70% (green and blue) respectively. While we have obtained good results from the first iteration of AP, allowing the GP to retrain with new data has shown us that further improvements can be made and thus we deemed 10 iterations as sufficient.

For the L2 circuit, since we do not know the true minimum and maximum values of the responses, we have opted to have a higher number of iterations (50), as seen in Table 1, AP 50 rows. Here, we can see the improvements brought in terms of absolute values. For GM the method manages to identify on OCC which leads to specification violations (e.g., GM = 8), while for PM and PSRR we do not see any significant improvements.

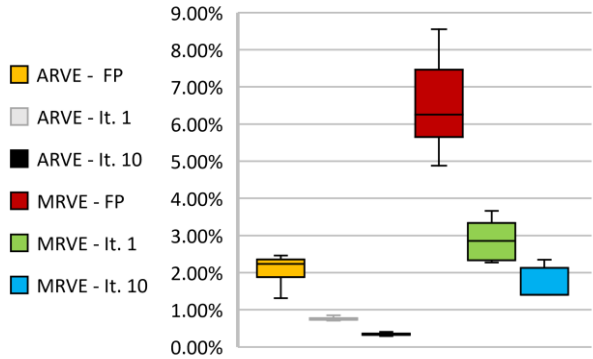

**Fig. 2** Baseline and updated results for synthetic circuit. We ran the experiment with 10 seeds in order to assure consistency.

**Table 1.** Baseline and updated results for L2

|  |  | Min. | Max. | Average |
|---|---|---|---|---|
| GM | FP | 9.28 | 10.26 | 9.80 |
|  | AP 1 | 10.06 | 19.91 | 8.99 |
|  | AP 50 | **8.00** | 8.99 | 8.37 |
| PM | FP | 28.64 | 32.34 | 30.81 |
|  | AP 1 | 35.13 | 54.52 | 40.14 |
|  | AP 50 | 28.64 | 29.52 | 28.97 |
| PSRR | FP | -33.79 | -33.64 | -33.72 |
|  | AP 1 | -38.65 | -36.10 | -33.60 |
|  | AP 10 | -33.61 | -33.60 | -33.60 |

## 5. Conclusion

In this paper we presented a novel adaptive search algorithm which can offer extended coverage in analog circuit verification. In addition to our prior work [7][8], the proposed algorithm also considers process variation. The algorithm was evaluated on both synthetic and real circuits, and we report that it is able to find a specification violation, as opposed to the baselines which miss to evaluate the circuit in those specific operating conditions. For future work, we plan to evaluate the algorithm on more circuits as well as to improve the candidate selection component.

**Acknowledgments.** This work was supported by a grant of the Ministry of Research, Innovation and Digitization, CCCDI - UEFISCDI, project number PN-III-P2-2.1-PTE-2021-0460, within PNCDI III.